\documentclass[letterpaper, 10 pt, conference]{ieeeconf}  
\usepackage{multirow}
\usepackage{times}
\usepackage{epsfig}
\usepackage{graphicx}
\usepackage{amsmath}
\usepackage{amssymb}
\usepackage{color}
\usepackage{diagbox}
\usepackage{interval}
\usepackage{amsmath}
\usepackage[table]{xcolor}
\usepackage{authblk}
\usepackage{etoolbox}
\makeatletter
\patchcmd{\@makecaption}
  {\scshape}
  {}
  {}
  {}
\makeatletter
\patchcmd{\@makecaption}
  {\\}
  {.\ }
  {}
  {}
\makeatother

\usepackage{tikz}

\IEEEoverridecommandlockouts                              

\title{\LARGE \bf
Beyond Photometric Loss for Self-Supervised Ego-Motion Estimation}

\author{Tianwei Shen$^1$, Zixin Luo$^1$, Lei Zhou$^1$, Hanyu Deng$^1$, Runze Zhang$^2$, Tian Fang$^3$, Long Quan$^1$ 
\thanks{$^{1}$Authors are with the Department of Computer Science and Engineering, Hong Kong University of Science and Technology {\tt\small \{tshenaa, zluoag, lzhouai, hdeng, quan\}@cse.ust.hk}}%
\thanks{$^{2}$Runze Zhang is with Tencent YouTu Lab, Shenzhen. {\tt\small ryanrzzhang@tencent.com}}%
\thanks{$^{3}$Tian Fang is with Everest Innovation Technology (Altizure), Hong Kong. {\tt\small fangtian@altizure.com}}
}

\begin{document}

\maketitle
\thispagestyle{empty}
\pagestyle{empty}

\begin{abstract}
Accurate relative pose is one of the key components in visual odometry (VO) and simultaneous localization and mapping (SLAM). Recently, the self-supervised learning framework that jointly optimizes the relative pose and target image depth has attracted the attention of the community. Previous works rely on the photometric error generated from depths and poses between adjacent frames, which contains large systematic error under realistic scenes due to reflective surfaces and occlusions. In this paper, we bridge the gap between geometric loss and photometric loss by introducing the matching loss constrained by epipolar geometry in a self-supervised framework. Evaluated on the KITTI dataset, our method outperforms the state-of-the-art unsupervised ego-motion estimation methods by a large margin. The code and data are available at \color{blue}{https://github.com/hlzz/DeepMatchVO}.
\end{abstract}

\section{Introduction}
Simultaneous localization and and mapping (SLAM) and visual odometry (VO) serve as the basis for many emerging technologies such as autonomous driving and virtual reality. Among various implementations that rely on different sensors, the monocular approach is advantageous in mobile robot with limited budgets. Although it is sometimes unstable compared with stereo inputs or fusing more sensors such as IMU and GPS, it is still desirable considering the low cost and applicability. The visual system of humans also serves as the the proof of existence for an accurate visual monocular SLAM system. We humans are capable of perceiving the environment even viewing a scene with one eye. Several monocular cues such as motion parallax~\cite{ferris1972motion} and optical expansion~\cite{swanston1986perceived} embed prior knowledge into depth sensing. Enlightened by the biological resemblance, the joint inference of depth and relative motion~\cite{zhou2017unsupervised,vijayanarasimhan2017sfm,yin2018geonet} has recently attracted the attention of the visual SLAM community. Given $N$-adjacent frames, this method uses CNN to predict the depth map of the target image and the relative motion of the target frame to other source frames. With depth and pose, the source image can be projected onto the target frame to synthesize the target view. It minimizes the error between the synthesis view and the actual image.

There are generally two sources of information that involve the interaction of depth and motion: photometric information like intensity and color from images~\cite{engel2017direct}, and geometric information computed from stable local keypoints~\cite{luo2018geodesc}. Most unsupervised or self-supervised methods for depth and motion estimation utilize image reconstruction error based on photometric consistency. Given known camera intrinsics, the approach would not require large amount of labelled data, making it more general and applicable to a broader ranges of applications. However, the unsupervised learning formulation enforces strong assumptions that require the scenes to be static without dynamic objects, the modeling surface to be Lambertian, and no occlusion exists between adjacent views. These criterions generally do not hold in a real-world scenario, even for a very short camera baseline. For example, the state-of-the-art single-view depth estimation result is obtained by training with 3 consecutive frames, but not on longer image sequences such as using 5 frames, as demonstrated in several previous works~\cite{zhou2017unsupervised,yin2018geonet}. This implies that photometric error would accumulate for wide baselines (5 rather than 3 frames), which further shows the limitation of using only photometric error as the supervision.

We show in this paper that the self-generated geometric quantities can be implicitly embedded into the training process without breaking the simplicity of inference. Specifically, we explore intermediate geometric information such as pairwise matching and weak geometry generated automatically to improve the joint optimization for depth and motion. These intermediate geometric representations are much less likely to be affected by the intrinsic photometric limitations. We also analyze the intrinsic flaw with per-pixel photometric error and propose a simple percentile mask to mitigate the problem. The method is evaluated on the KITTI dataset, which achieves the best relative pose estimation performance of its kind. In addition, we demonstrate a VO system that chains and averages the predicted relative motions for full trajectory, which even outperforms monocular ORB-SLAM2 without loop closure on KITTI Odometry Sequence 09.

\section{Related Works}
\begin{figure*}[th]
	\centering 
	\includegraphics[width=\textwidth]{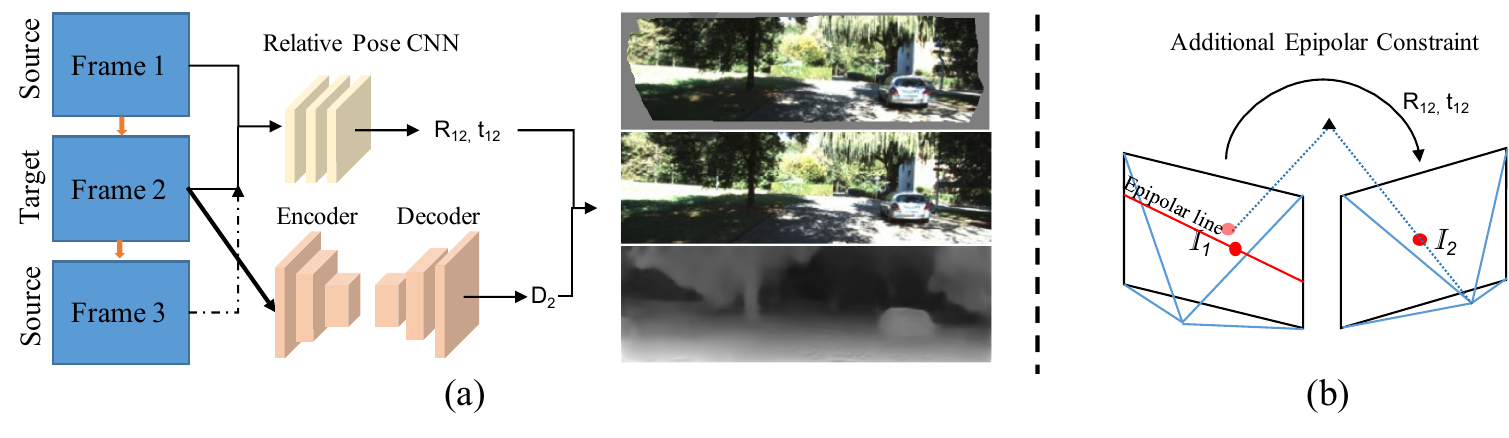}
	\caption{Architecture of our method. (a) The proposed method takes several adjacent frames as input and output the depth image of the target image and relative poses, abided by both photometric loss and geometric loss. (b) The geometric constraint is enforced by pairwise matching and epipolar geometry.}
	\label{fig:arch}
\end{figure*}

In this section, we discuss the related works on traditional visual VO/SLAM systems and learning-based methods for visual odometry. 

\subsection{Traditional visual SLAM approaches} 
Current state-of-the-art visual SLAM approaches can be generally characterized into two categories: indirect and direct formulations. Indirect methods conquer the motion estimation problem by first computing some stable and intermediate geometric representations such as keypoint~\cite{mur2017orb}, edgelet~\cite{klein2008improving} and optical flow~\cite{ranftl2016dense}. \textit{Geometric} error is then minimized using these reliable geometric representations either with sliding-window or global bundle adjustment~\cite{triggs1999bundle}. This is the most widely-used formulation for SLAM systems~\cite{davison2007monoslam,konolige2008frameslam,mur2017orb}.

For visual odometry or visual SLAM (vSLAM), direct methods directly optimize the \textit{photometric} error which corresponds to the light value received by the actual sensors. Examples include ~\cite{newcombe2011dtam,engel2014lsd,engel2017direct}. Given accurate photometric calibration information (such as gamma correction, lens attenuation), this formulation spares the costly sparse geometric computation and could potentially generate finer-grained geometry like per-pixel depth. However, this formulation is less robust than indirect ones with the presence of dynamic moving objects, reflective surfaces and inaccurate photometric calibration. Note that the self-supervised learning framework derives from the direct method.

\subsection{Learning Depth and Pose from Data}
Most of pioneering depth estimation works rely on supervision from depth sensors~\cite{saxena2006learning,eigen2014depth}. Ummenhofer et al.~\cite{ummenhofer2017demon} propose an iterative supervised approach to jointly estimate optical flow, depth and motion. This iterative process allows the use of stereopsis and gives fairly good results given depth and motion supervision. 

The self-supervised approaches for structure and motion borrow ideas from warping-based view synthesis~\cite{zitnick2004high}, a classical paradigm of which is to composite novel view based on the underlying 3D geometry. 
Garg et al.~\cite{garg2016unsupervised} propose to learn depth using stereo camera pairs with known relative pose. Godard et al.~\cite{godard2017unsupervised} also rely on calibrated stereo to obtain monocular depth with left-right consistency checking. Zhan et al.~\cite{zhan2018unsupervised} consider deep features from the neural nets in addition to the photometric error. The above three methods have limited usability in the monocular scenario where the pose is unknown.
Zhou et al.~\cite{zhou2017unsupervised} and Vijayanarasimhan et al.~\cite{vijayanarasimhan2017sfm} develop similar joint learning methods for the traditional structure and motion (SfM) problem~\cite{shen2016graph,zhu2018very}, with the major difference that~\cite{vijayanarasimhan2017sfm} can incorporate supervised information and directly solve for dynamic object motion. 
Later, ~\cite{wang2018learning} discuss the critical scale ambiguity issue for monocular depth estimation, which is neglected by previous works. To resolve scale ambiguity, the estimated depth is first normalized before being fed into the loss layer. Geometric constraints of the scene are enforced by an approximate ICP based matching loss in~\cite{mahjourian2018unsupervised}. For real-world applications, pose and depth estimation using CNNs have also been integrated into visual odometry systems~\cite{wang2017deepvo,li2017undeepvo}. Ma et al.~\cite{ma2017sparse} consider the sparse depth measurements with RGB data to reconstruct the full depth map.

The above view-synthesis-based methods~\cite{zhou2017unsupervised,vijayanarasimhan2017sfm,li2017undeepvo,mahjourian2018unsupervised} is based on the assumptions that the modeling scene is static and the camera is carefully calibrated to get rid of photometric distortions such as automatic exposure changes and lens attenuation (vignetting)~\cite{kim2008robust}.
This problem becomes serious as most of the previous works train models on KITTI~\cite{Geiger2013IJRR} or Cityscapes~\cite{cordts2016cityscapes} datasets, in which the camera calibration does not consider non-linear response functions (gamma-correction / white-balancing) and vignetting. As the input image size is limited by the GPU memory, the pixel value information is further degraded by down-sampling. 

These learning-based methods optimizing \textit{photometric error} corresponds to the direct methods~\cite{engel2014lsd,engel2017direct} for SLAM. Indirect methods~\cite{davison2007monoslam,mur2017orb}, on the other hand, decompose the structure and motion estimation problem by first generating an intermediate representation and then computing the desired quantities based on \textit{geometric loss}. These intermediate representations like keypoints~\cite{luo2018geodesc,rublee2011orb} are typically stable and resilient to occlusions and photometric distortions. In this paper, we advocate to import geometric losses into the self-supervised depth and relative pose estimation problem.  

\section{Methods}
\subsection{Overview}
Our method combines the accurate intermediate geometric representations of traditional monocular SLAM with self-supervised depth estimation to deliver a better formulation for joint depth and motion estimation. Figure~\ref{fig:arch} shows the architecture of our method with concatenated three adjacent frames $(\mathcal{I}_1, \mathcal{I}_2, \mathcal{I}_3)$ as input, and the predicted depth map of the target frame and relative poses as output. We first give a brief overview of previous works that rely much on the photometric errors.

Taken two adjacent frames $\mathcal{I}_1$ and $\mathcal{I}_2$ as an example (the case for frame $\mathcal{I}_3$ and $\mathcal{I}_2$ is the same), the pose module takes the concatenated image and output a 6-DoF relative pose  $[\hat{R_{12}} | \hat{t_{12}}]$ in an end-to-end fashion. The depth module, which is a encoder-decoder network, takes the target frame $\mathcal{I}_2$ as input to generate the depth map for $\mathcal{I}_2$, denoted as $\widehat{D_2}$. 

The typical methods~\cite{garg2016unsupervised,godard2017unsupervised,zhou2017unsupervised,vijayanarasimhan2017sfm,yin2018geonet,mahjourian2018unsupervised,zhan2018unsupervised,wang2018learning} for unsupervised estimation of $\widehat{D_2}$ and $(\hat{R_{12}}, \hat{t_{12}})$ are to employ the image synthesis loss. Suppose $p_2$ denotes a pixel in $\mathcal{I}_2$ that is also visible in $\mathcal{I}_1$, its projection $p_1$ on $\mathcal{I}_1$ is represented by
\begin{equation}\label{eqn1}
p_1 \sim K_1 [\hat{R_{12}} | \hat{t_{12}}]  \widehat{D_2}(p_2) K_2^{-1} p_2
\end{equation}
where $\sim$ mean `equal in the homogeneous coordinate' and $K_1$ and $K_2$ are the intrinsic matrix for the corresponding two images. Given this relation, we can obtain a synthesis image $\widetilde{\mathcal{I}_2^1}$ using source frames $\mathcal{I}_1$ by bilinear sampling~\cite{jaderberg2015spatial}. Depth and relative pose are then optimized by the image reconstruction loss between $\widetilde{\mathcal{I}_2^1}$ and $\mathcal{I}_2$. Early works usually adopt the $L1$ loss of the corresponding pixels while later structured similarity~\cite{wang2004image} (SSIM) is introduced to evaluate the quality of image predictions. We follow~\cite{yin2018geonet,mahjourian2018unsupervised} among others and use the combination of the both $L1$ loss and SSIM loss as the image reconstruction loss $\mathcal{L}_{img}$:
\begin{equation}
\mathcal{L}_{img} = (1-\alpha) ||\mathcal{I}_2 -  \widetilde{\mathcal{I}_2^1} ||_1 + \alpha \frac{1-SSIM(\mathcal{I}_2 -  \widetilde{\mathcal{I}_2^1})}{2}
\end{equation}
where $\alpha$ is the balancing factor which we set to 0.85~\cite{yin2018geonet,mahjourian2018unsupervised}. This loss formulation should be accompanied with a smoothness term to resolve the gradient-locality issue in motion estimation~\cite{bergen1992hierarchical} and remove discontinuity of the learned depth in low-texture regions. We adopt the edge-aware depth smoothness loss in ~\cite{yin2018geonet} which uses image gradient to weight depth gradient:
\begin{equation}
\mathcal{L}_{smooth} = \sum_{p} |\nabla D(p)|^T \cdot e^{-|\nabla I(p)|}
\end{equation}
where $p$ is the pixel on the depth map $D$ and image $I$, $\nabla$ denotes the 2D differential operator, and $|\cdot|$ is the element-wise absolute value.

\subsection{Geometric Error from Epipolar Geometry}
One of the main reasons for the success of indirect SLAM method is the use of stable invariants computed from raw image input, such as keypoints and line segments. Though still computed from pixel values, descriptors for these stable image patches have strong invariance guaranteed by scale-space theory~\cite{lindeberg1994scale}. For learning-based approaches, these geometric ingredients can be pre-processed offline and implicitly integrated into CNNs. In this paper, we demonstrate the boost of several geometric elements to overcome the intrinsic drawbacks of the current approaches. 

One of the fundamental building blocks for sparse-feature-based SLAM or SfM is the pairwise matching with geometric verification. For a pair of overlap images ($\mathcal{I}_1, \mathcal{I}_2$) viewing the same scene with canonical relative motion $(R_{12}, t_{12})$, a set of feature matches $\{(p_{i}, q_{i})\}$ in the homogeneous image coordinates can be reliably obtained. Then the following epipolar geometry constraint holds:
\begin{equation}
q_{i}^{T}F_{12}p_{i} = (K_2^{-1}q_{i}')^T  R_{12}[t_{12}]_\times (K_1^{-1}p_{i}') = 0
\end{equation}
where $F_{12}$ is the corresponding fundamental matrix, $p_{i}'$ and $q_{i}'$ represent the homogeneous camera coordinates of the $i$-th matched points, and $K_1$ and $K_2$ are their corresponding intrinsic matrix. $[\cdot]_\times$ is the matrix representation of the cross product with a 3-dimensional vector.

We use the projection error from the first image to the second image as the supervision signal for relative pose estimation. $l_{12}^{(i)} = F_{12}p_{i}$ defines the epipolar line~\cite{hartley2003multiple} on which $q_{i}$ must lie on. Therefore, the geometric loss $\mathcal{L}_{geo}$ is defined by the sum of the distance from point to line for all (or sampled) corresponding matches.
\begin{equation}
\mathcal{L}_{geo} = \sum_{i} dist(l_{12}^{(i)}, q_{i})
\end{equation}
where the 2D point $(x_0, y_0)$ to line $ax+by+c=0$ distance is defined by $dist(ax+by+c=0, (x_0, y_0)) = \frac{|ax_0+by_0+c |}{\sqrt{a^2+b^2}}$, and the sum iterates over corresponding image matches in adjacent frames.

\begin{figure*}[th]
	\centering 
			\resizebox{0.97\textwidth}{!}{ 
	\includegraphics[width=\textwidth]{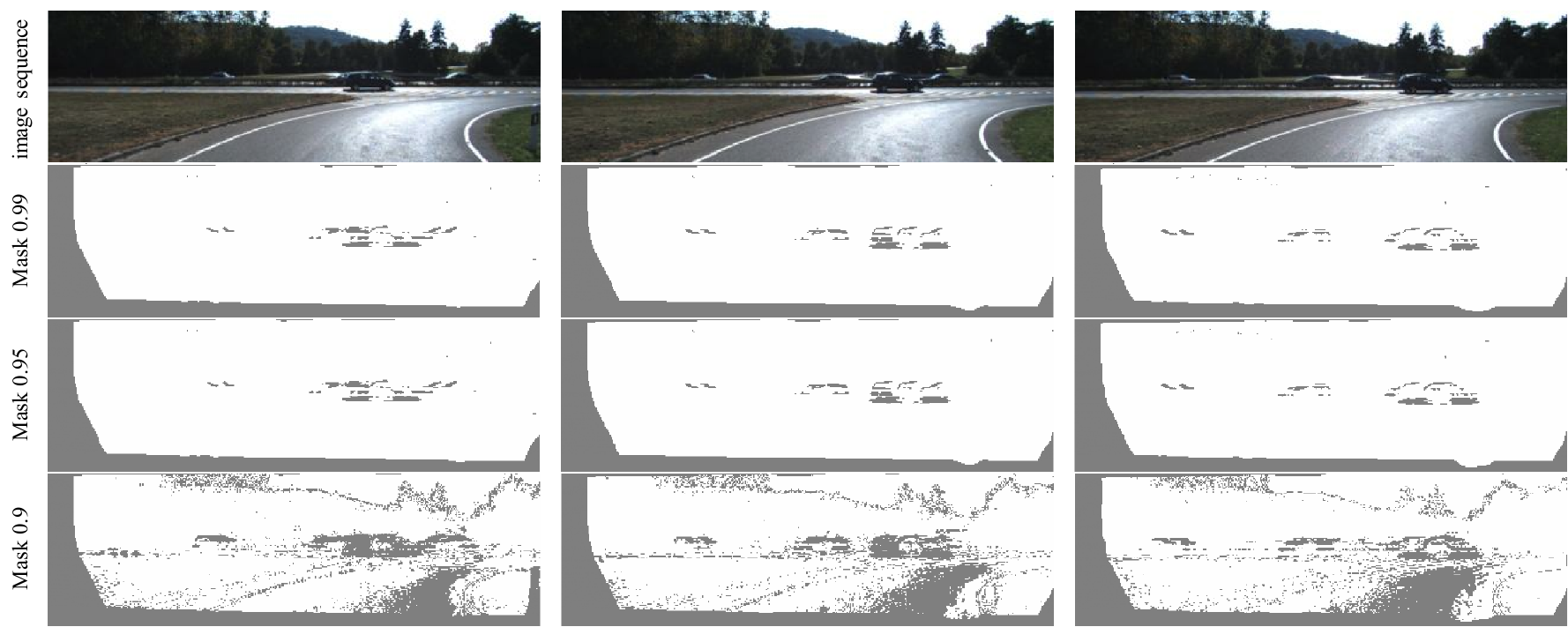}
}
	\caption{Threshold masks with $P_M$ from 0.9 to 0.99. We set $P_M(=0.99)$ modestly so that the loss formulation does not lose too much information.}
	\label{fig:thresh_mask}
\end{figure*}

\subsection{Other weak pairwise geometric supervisions} \label{sec:other_geo}
To incorporate geometric losses into the self-supervised framework, several intermediate geometric computations can be employed. Apart from using epipolar geometry (\textit{Pairwise-Matching}), indirect methods have provided other forms of geometric supervisions, such as the camera pose computed using perspective-n-point (PnP) algorithms~\cite{gao2003complete,lepetit2009epnp}. Since these properties can be computed offline, it belongs to the self-supervised category to utilize the weak geometric supervisions.
With 3d point to 2d projection matches, we can obtain a set of inaccurate/weak supervision for absolution camera poses $\{P_i = (R_i, T_i)\}$ for $\{\mathcal{I}_i\}$. We have explored two ways of incorporating the weak supervision. The first one, denoted as \textit{Direct-Weak-Pose}, is to directly use the weak poses without explicitly learning the relative pose CNN. Since the weak poses are absolute with respect to the current scene (instead of the relative ones learned from the pose CNN), Equation (\ref{eqn1}) becomes
\begin{equation}\label{eqn:abs_pose}
\begin{split}
p_1 & \sim K_1  P_1 P_2^{-1}\widehat{D_2}(p_2) K_2^{-1} p_2 \\
& \sim K_1 [R_1 | T_1] [R_2^T | -R_2^T T_2] \widehat{D_2}(p_2) K_2^{-1} p_2\\
& \sim K_1 [R_1R_2^T | T_1 - R_1R_2^T T_2] \widehat{D_2}(p_2) K_2^{-1} p_2
\end{split}
\end{equation}

The second way is to use the weak pose as a prior~\cite{klodt2018supervising}, which we denote as \textit{Prior-Weak-Pose}. Different from \textit{Direct-Weak-Pose}, the pose CNN is used for relative pose estimation, while its deviation from the weak pose computed using traditional geometric methods is added to the optimization. Formally, \textit{Prior-Weak-Pose} considers one additional prior pose loss written as 
\begin{equation}
\begin{split}
\mathcal{L}_{pose} &= \mathcal{L}_{rotation} + \mathcal{L}_{translation}\\
&= w_{r} ||\hat{r_{ij}} - \overline{r_{ij}} ||_2 + w_t ||\hat{t_{ij}} - \overline{t_{ij}} ||_2
\end{split}
\end{equation}
where $(\hat{r_{ij}}, \hat{t_{ij}})$ and $(\overline{r_{ij}}, \overline{t_{ij}})$ are the estimated 6-DoF relative motion and weak pose, with rotation in Euler angle form and translation normalized $||\hat{t_{ij}} ||_2  = ||\overline{t_{ij}}||_2  = 1$. $w_r$ and $w_t$ are weights for rotation part and translation part respectively. Yet, we will show that the both ways of using weak poses computed from traditional methods like~\cite{mur2017orb} are worse than the proposed method that utilizes raw feature matches.

\subsection{Fixing the Photometric Error}\label{sec:thr_mask}
As photometric error is inevitably one of the major supervision signals, we also consider mitigating the systematic error rooted in the optimization process. To this end, we introduce a simple solution that works well in practice. Since occlusions and dynamic objects prevalently exist in images, previous work such as~\cite{zhou2017unsupervised,vijayanarasimhan2017sfm} further train a network to mask out these erroneous regions. Yet, this approach only brings marginal performance boost because it entangles with the depth and motion networks. Instead of learning the mask, we propose a deterministic mask that is computed on-the-fly. During the training process, we compute the mask $\mathcal{M}(P_M)$ based on the distribution of image reconstruction loss, defined as 
\begin{equation}
\begin{array}{cc}
\mathcal{M}(P_M) =   \{ & 
    \begin{array}{cc}
      1 & Percentile(\mathcal{L}_{img}(i,j)) \leq P_M \\
      0 & otherwise
    \end{array}
\end{array}
\end{equation}
where pixel positions $(i,j)$ whose photometric loss is above a percentile threshold $P_M$ are filtered out. This is based on the fact that objects or regions that do not obey the static assumption usually impose larger errors. Throughout the experiment, we fix $P_M$ to be 0.99 which is a modest choice that filters out extremely false regions while preserves much of the image content to facilitate the optimization (as shown in Figure~\ref{fig:thresh_mask}.). Experiments validate that this simple strategy improves the depth estimation by better handling occlusions and reflections.

In the end, the total loss becomes 
\begin{equation}
\mathcal{L}_{total} = \mathcal{M}(P_M) \odot \mathcal{L}_{img} + w_s \mathcal{L}_{smooth} + [w_g \mathcal{L}_{geo}] + [w_p \mathcal{L}_{pose} ]
\end{equation}
where $w_s, w_g, w_p$ are weights for the smoothness loss, geometric loss and weak pose loss respectively. The smoothness weight $w_s$ is set to 0.1 throughout the evaluation. As for $w_g$ and $w_p$, since the weak geometric prior used in $\mathcal{L}_{pose}$ has the same functionality as the pairwise matching used for $\mathcal{L}_{geo}$, we add the two losses separately and compare their performance. We refer to the case where $(w_g = 0.001, w_p = 0)$ as the \textit{Pairwise-Matching} approach, while the case where $(w_g= 0, w_p = 0.1)$ as the \textit{Prior-Weak-Pose} approach. As we describe in Section~\ref{sec:other_geo}, we can also directly use the pose computed from PnP algorithms as the pose supervision. In this case (\textit{Direct-Weak-Pose}), we only train the depth network for monocular depth estimation with $(w_g = 0, w_p = 0)$. The performance comparison for these three approaches is shown in the ablation study.

\section{Experiments}
\begin{table*}[]
\centering
\caption{Three ways to incorporate geometric constraints, compared with baseline method with and without mask. The columns that are marked with red means `the lower the better', and the columns with purple means `the higher the better'. }
\resizebox{\textwidth}{!}{ 
\begin{tabular}{|l|c|c|c|c|c|c||c|c|c|}
\hline
Method              & Geometric Info   & Cap (m) & \cellcolor{red!25}Abs Rel         & \cellcolor{red!25} Sq Rel &  \cellcolor{red!25}RMSE   &  \cellcolor{red!25}RMSE log &  \cellcolor{blue!25}$\delta <1.25 $ &  \cellcolor{blue!25}$\delta < 1.25^2 $& \cellcolor{blue!25} $\delta < 1.25^3$            \\ \hline
\textit{Baseline (\textbf{w/o} Mask)}   & No             & 80      &    0.171         &  1.512   &  6.332  & 0.250   &    0.764        &      0.918      &    0.967                    \\ \hline
\textit{Baseline (\textbf{w} Mask)}   & No             & 80      &    0.163         &  1.370   &  6.397  & 0.258     &    0.758        &      0.910      &    0.962                    \\ \hline
\textit{Pairwise-Matching}   & Self-generated Matches             & 80      &    \textbf{0.156}     & \textbf{1.357}  & \textbf{6.139}  &   \textbf{0.247}  &  \textbf{0.781}         &      \textbf{0.920}     &        0.965               \\ \hline
\textit{Prior-Weak-Pose}~\cite{klodt2018supervising}       & Self-generated Pose               & 80      &    0.163        & 1.371  & 6.275  & 0.249    &   0.773         &   0.918         &  \textbf{0.967}                     \\ \hline
\textit{Direct-Weak-Pose} & Self-generated Pose                 & 80      &     0.162      & 1.46& 6.27 & 0.249   &  0.775          &  0.919          &    0.965                  \\ \hline
\end{tabular}\label{tab:ablation}
}
\end{table*}

\subsection{Dataset}
\emph{KITTI.} We evaluate our method on the most common KITTI~\cite{Geiger2013IJRR,Menze2015CVPR} benchmark dataset, which includes a full set of input sources including raw images, 3D point cloud data from LIDAR and camera trajectories. To conduct a fair comparison with related works, we adopt the Eigen split for single-view depth benchmark and use the odometry sequences to evaluate the visual odometry performance. All the training and testing images are from the left monocular camera from the stereo pair and down-sampled to $128\times416$.

\emph{Eigen Split.} We evaluate the single-view depth estimation performance on the test split composed of 697 images from 28 scenes as in~\cite{eigen2014depth}. Images in the test scenes are excluded in the training set. Since the test scenes overlaps with the KITTI odometry split (i.e. some test images of Eigen split are contained in the KITTI odometry training set, and vice versa), we train the model solely on the Eigen split with 20129 training images and 2214 validation images.

\emph{KITTI odometry.} The KITTI odometry dataset contains 11 driving sequences with ground-truth poses and depth available (and 11 sequences without ground-truth). For pose estimation, we train the model on KITTI odometry sequence 00-08 and evaluate the pose error on sequence 09 and 10. 18361 images are used for training and 2030 for validation.

\emph{Cityscapes.} We also try pre-training the model on the Cityscapes~\cite{cordts2016cityscapes} dataset too boost performance. The process is conducted without geometric loss for 60k steps, with 88084 images for training and 9659 images for validation.

\subsection{Implementation Details}
\emph{Geometric Supervision.} We use SIFT descriptor for feature matching~\cite{zhou2017progressive}, which is widely used for SfM. The average feature number for each image is around 8000. For weakly-supervised poses, we use the consecutive motion generated by PnP algorithm used in stereo ORB-SLAM2~\cite{mur2017orb}, which is essentially EPnP~\cite{lepetit2009epnp} with RANSAC~\cite{fischler1981random}. We choose the stereo version rather than the monocular one because (1) it is more accurate than monocular (but still cannot be viewed as the ground truth) and (2) the initialization process takes the initial stereo pair and all frames get reconstructed, while the first few frames may be missing in the monocular version.
For feature matching supervision, pairwise matching is conducted between adjacent frames filtered by the epipolar geometry using the normalized eight-point algorithm~\cite{hartley1997defense}, which leads to around 2000 fundamental matrix inliers for adjacent frames. We random sample 100 matching features of each image pair for training.

\emph{Learning.} We implement the neural nets using the Tensorflow~\cite{abadi2016tensorflow} framework. During training, we use the Adam~\cite{kingma2014adam} solver with $\beta_1 = 0.9$, $\beta_2 = 0.999$, a learning rate of 0.0001 and a batch size of 4. We use ResNet-50~\cite{he2016deep} as the depth encoder and the same architecture for pose network as~\cite{zhou2017unsupervised}. Most of the training tasks usually converge within 200k iterations.
To address the gradient locality issue, many works~\cite{zhou2017unsupervised,yin2018geonet} take the multi-scale approach to allow gradients to be derived from larger spatial regions. As this approach alleviates the problem a bit, it also brings new error since low-resolution images have inaccurate photometric values. We therefore only use one image scale for training without down-sampling, and observe a slight improvement for depth estimation performance. 

\subsection{Ablation Study}
We first show that adding the threshold mask (section~\ref{sec:thr_mask}) improves the depth estimation (the first two items of Table~\ref{tab:ablation}), and then compare three ways of incorporating geometric information, namely \textit{Pairwise-Matching}, \textit{Prior-Weak-Pose}~\cite{klodt2018supervising} and \textit{Direct-Weak-Pose}. Since pose data is more conveniently generated from the KITTI odometry sequences, we train the models on KITTI odometry sequence 00-08 and evaluate the monocular depth estimation performance on the Eigen split test set. Since some test images in the Eigen split test set are included in the training sequence 00-08, we remove the in-training test samples using matchable image retrieval~\cite{shen2018matchable}. Therefore, the result is not comparable with Table~\ref{tab:depth_est} because the test sets are different.
Note that here we do not directly compare the pose estimation performance because \textit{Direct-Weak-Pose} does not even learn to estimate pose. The error measures conform with the one used in ~\cite{eigen2014depth}. 
\begin{equation*}
\resizebox{0.48\textwidth}{!}{$
Abs\ Rel: \frac{1}{|\mathcal{I}|} \sum_{ \mathcal{I}} \frac{|d_{ij}^{pred} - d_{ij}^{gt}|}{d_{ij}^{gt}} \  Sq\ Rel: \frac{1}{|\mathcal{I}|} \sum_{ \mathcal{I}} \frac{||d_{ij}^{pred} - d_{ij}^{gt}||}{d_{ij}^{gt}}
$}
\end{equation*}
\begin{equation*}
\resizebox{0.5\textwidth}{!}{$
RMSE: \sqrt{\frac{1}{|\mathcal{I}|} \sum_{\mathcal{I}} ||d_{ij}^{pred} - d_{ij}^{gt}||^2}  \  
RMSE\ log: \sqrt{\frac{1}{|\mathcal{I}|} \sum_{\mathcal{I}} ||\log{d_{ij}^{pred}} - \log{d_{ij}^{gt}}||^2} 
$}
\end{equation*}
\begin{equation*}
\resizebox{0.5\textwidth}{!}{$
Accuracy: \text{percent of } d^{pred}_{ij} s.t. \max (\frac{d_{ij}^{pred}}{d_{ij}^{gt} }, \frac{d_{ij}^{gt}}{d_{ij}^{pred} }) = \delta < 1.25, 1.25^2, 1.25^3
$}
\end{equation*}
where $|\mathcal{I}|$ is the total number of pixels in image $\mathcal{I}$.
As shown in Table~\ref{tab:ablation}, \textit{Pairwise-Matching} achieves the best depth estimation performance among the three. This is explainable because \textit{Prior-Weak-Pose} and \textit{Direct-Weak-Pose} both introduce the geometric bias in the estimation algorithms, while \textit{Pairwise-Matching} uses the raw matches.

\subsection{Depth Estimation}
\begin{table*}[]
\centering
\caption{Single-view depth estimation performance. The statistics for the compared methods are excerpted from the original papers. `K' represents KITTI raw dataset (Eigen split) and CS represents cityscapes training dataset. The best results with capped 80m are \textbf{bolded}.}
\resizebox{\textwidth}{!}{ 

\begin{tabular}{|l|c|c|c|c|c|c|c||c|c|c|}
\hline
Method              & Supervision   & Dataset & Cap (m) & \cellcolor{red!25}Abs Rel         &\cellcolor{red!25} Sq Rel & \cellcolor{red!25}RMSE   & \cellcolor{red!25}RMSE log &\cellcolor{blue!25} $\delta <1.25 $ &\cellcolor{blue!25} $\delta < 1.25^2 $& \cellcolor{blue!25} $\delta < 1.25^3$            \\ \hline
Eigen et al.~\cite{eigen2014depth} Fine   & Depth         & K       & 80      & 0.203           & 1.548  & 6.307  & 0.282    & 0.702           & 0.890           & 0.958                      \\ \hline
Liu et al.~\cite{liu2016learning}          & Depth         & K       & 80      & 0.202           & 1.614  & 6.523  & 0.275    & 0.678           & 0.895           & 0.965                      \\ \hline
Godard et al.~\cite{godard2017unsupervised}       & Pose          & K       & 80      & \textbf{0.148}           & 1.344  & 5.927  & 0.247    & 0.803           & 0.922           & 0.964 \\ \hline
Zhou et al.~\cite{zhou2017unsupervised} updated & No            & K       & 80      & 0.183           & 1.595  & 6.709  & 0.270    & 0.734           & 0.902           & 0.959                      \\ \hline
Mahjourian et al.~\cite{mahjourian2018unsupervised}   & No            & K       & 80      & 0.163           & 1.24   & 6.22   & 0.25     & 0.762           & 0.916           & 0.968                      \\ \hline

Yin et al. ~\cite{yin2018geonet}         & No            & K       & 80      & 0.155           & 1.296  & 5.857  & 0.233    & 0.793           & 0.931           & \textbf{0.973} \\ 
Yin et al. ~\cite{yin2018geonet}         & No            & K + CS       & 80      & 0.153           & 1.328  & 5.737  & 0.232    & \textbf{0.802}           & 0.934           & 0.972 \\ \hline
Ours                & No            & K       & 80      & 0.156 & 1.309 & 5.73 & 0.236   & 0.797          & 0.929          & 0.969                     \\ 
Ours                & No            & K + CS       & 80      & 0.152 & \textbf{1.205} & \textbf{5.564} & \textbf{0.227}   & 0.8          & \textbf{0.935}          & \textbf{0.973}                     \\ \hline \hline
Garg et al.~\cite{garg2016unsupervised}         & Stereo (Pose) & K       & 50      & 0.169           & 1.080  & 5.104  & 0.273    & 0.740           & 0.904           & 0.962                      \\ \hline
Zhou et al.~\cite{zhou2017unsupervised} & No            & K       & 50      & 0.201           & 1.391  & 5.181  & 0.264    & 0.696           & 0.900           & 0.966                      \\ \hline
Ours                & No            & K       & 50      & 0.149 & 1.01 & 4.36 & 0.222   & 0.812          & 0.937          & 0.973                     \\ \hline

\end{tabular}\label{tab:depth_est}
}
\end{table*}
Further, we compare our model trained with pairwise matching loss (\textit{Pairwise-Matching}) on KITTI Eigen train/val split with various approaches with either depth supervision, pose supervision or no supervision (self-supervision). The evaluation process is similar to ~\cite{zhou2017unsupervised} and we match medians of the predicted depth and ground-truth depth since the predicted monocular depth is defined up to scale. All ground-truth depth maps are capped at 80m (maximum depth is 80m) except~\cite{garg2016unsupervised} that are capped at 50m.
As shown in Table~\ref{tab:depth_est}, match loss improves the baseline self-supervised approach~\cite{zhou2017unsupervised} by a large margin and achieves state-of-the-art performance compared with methods using sophisticated information such as optical flow~\cite{yin2018geonet} or ICP~\cite{mahjourian2018unsupervised}.

\subsection{Visual Odometry Performance}

\begin{table}[]
\centering
\caption{Visual odometry performance. Learning-based methods use
$128\times416$ images while ORB-SLAM2 uses original images. The pose snippet data is not available for ~\cite{mahjourian2018unsupervised} so it is not compared for full pose.}
\resizebox{0.48\textwidth}{!}{ 
\begin{tabular}{|l|c|c|c|c|}
\hline
\multirow{2}{*}{Method}               & \multicolumn{2}{l|}{Seq.09}     & \multicolumn{2}{l|}{Seq.10 (no loop)}     \\ \cline{2-5} 
                                      & Snippet                  & Full (m) & Snippet                  & Full (m) \\ \hline
ORB-SLAM2 (full, w LC)                       & 0.014 $\pm$ 0.008            &  7.08    & 0.012 $\pm$ 0.011            &   5.74   \\ \hline
ORB-SLAM2 (full, w/o LC)                       & -            &   38.56   & -          &   5.74   \\ \hline
Zhou et al.~\cite{zhou2017unsupervised} updated (5-frame)         & 0.016 $\pm$ 0.009            &  41.79    & 0.013 $\pm$ 0.009            &    23.78  \\ \hline
Yin et al. ~\cite{yin2018geonet} (5-frame)         & 0.012 $\pm$ 0.007            &  152.68    & 0.012 $\pm$ 0.009            &  48.19    \\ \hline
Mahjourian et al.~\cite{mahjourian2018unsupervised} , no ICP (3-frame)   & 0.014 $\pm$ 0.010            &   -   & 0.013 $\pm$ 0.011            &   -   \\ \hline
Mahjourian et al.~\cite{mahjourian2018unsupervised} , with ICP (3-frame) & 0.013 $\pm$ 0.010            &  -    & 0.012 $\pm$ 0.011            &    -  \\ \hline
Ours et al. (3-frame)                 & \textbf{0.0089 $\pm$ 0.0054} &  18.36    & \textbf{0.0084 $\pm$ 0.0071} &  16.15    \\ \hline
\end{tabular}\label{tab:pose_est}
}
\end{table}

\begin{figure}[th]
	\centering 
		\resizebox{0.45\textwidth}{!}{ 

	\includegraphics[width=\textwidth]{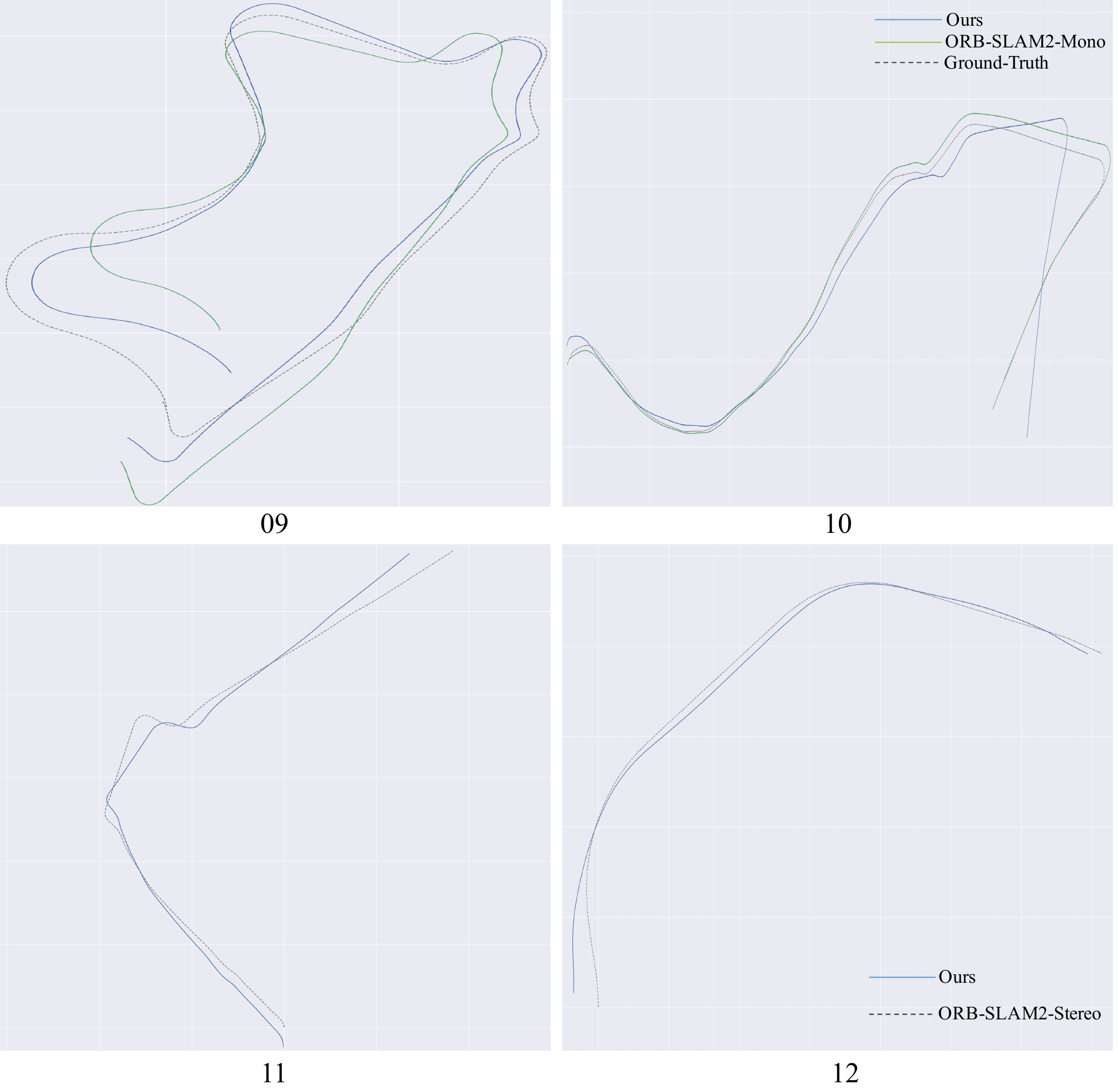}
	}
	\caption{KITTI sequence 09 and 10 trajectories. Our end-to-end model, monocular ORB-SLAM2 without loop closure, and ground-truth trajectories are shown (best view in color).}

	\label{fig:kitti_traj}
	
\end{figure}

Relative pose estimation is evaluated on the KITTI odometry sequence 09/10 and compared with both learning-based methods and traditional ones such as ORB-SLAM2~\cite{rublee2011orb}. Compared with depth estimation, we care much more about the relative pose estimation ability since the match loss directly interacts with it. We have observed that with the pairwise matching supervision, the result for visual odometry has been extensively improved. We first measure the Absolute Trajectory Error (ATE) over $N$-frame snippets ($N$=3 or 5), as measured in ~\cite{zhou2017unsupervised,yin2018geonet,mahjourian2018unsupervised}.
As shown in Table~\ref{tab:pose_est} (`Snippet' column), our method outperforms other state-of-the-art approaches by a large margin. 

However, simply comparing ATE over snippets is advantageous to the learning-based methods, since traditional methods like ORB-SLAM2 utilize window-based optimization over a longer sequence. Therefore, we chain the relative motions given by $N$-frames and apply simple motion averaging to obtain the full trajectory (1591 for seq.09 and 1201 for seq.10). The full pose is compared with the full trajectory computed by monocular ORB-SLAM2 approach without loop closure. Since the relative motion recovered by monocular visual odometry systems has an undefined scale, we first align the trajectories with the ground-truth using a similarity transformation from the evaluation package \textit{evo}\footnote{https://github.com/MichaelGrupp/evo}.

As shown in Table~\ref{tab:pose_est} (`Full' represents the median translation error measured in meters), our method has the lowest full trajectory error compared with similar methods due to the geometric supervision. Compared with ORB-SLAM2, our method achieves lower median ATE error (18.36m) on sequence 09 but is worse on sequence 10 (16.15m). Note that sequence 09 has a loop structure while sequence 10 does not, as shown in Figure~\ref{fig:kitti_traj}. We also show the trajectories of sequence 11 and 12 where the ground-truth poses are unavailable, using stereo ORB-SLAM2 results as the reference.
It is observed that the learned model has worse performance for rotation with large angles. This may be due to the lack of rotating motion in the KITTI training data as forward motion dominates the car movement. Considering the input smaller image size ($128\times 416$) and the simplicity of the implementation, this end-to-end visual odometry method still has great potential for future improvement.

\section{Conclusions}
In this paper, we first analyze the limitation of the previous loss formulation used for self-supervised depth and motion estimation. We then propose to incorporate intermediate geometric computations such as feature matches into the motion estimation problem. This paper is a preliminary exploration for the usability of geometric quantities in self-supervised motion learning problem. Currently, we only consider two-view geometric relations. Future directions include fusing geometric quantities in longer image sequences as in bundle adjustment~\cite{triggs1999bundle} and combing learning methods with traditional approaches as used in~\cite{tang2018ba,yang2018deep}.

\noindent\textbf{Acknowledgement.} This work is supported by Hong Kong RGC T22-603/15N and Hong Kong ITC PSKL12EG02. We also thank the generous support of Google Cloud Platform.

{\small
\bibliographystyle{ieee}
\bibliography{icra}
}

\end{document}